\documentclass{article}

\PassOptionsToPackage{numbers,compress}{natbib}
\usepackage[preprint]{neurips_2026}
\usepackage[utf8]{inputenc}
\usepackage[T1]{fontenc}
\usepackage{hyperref}
\usepackage{url}
\usepackage{booktabs}
\usepackage{amsfonts}
\usepackage{amsmath}
\usepackage{amssymb}
\usepackage{microtype}
\usepackage{xcolor}
\usepackage{graphicx}
\usepackage{array}
\usepackage{tabularx}
\usepackage{enumitem}
\usepackage{listings}
\usepackage{xspace}

\definecolor{darkblue}{RGB}{0,70,127}
\definecolor{darkgreen}{RGB}{0,120,50}
\definecolor{darkred}{RGB}{180,0,0}
\definecolor{lightgray}{RGB}{240,240,240}

\hypersetup{
  colorlinks=true,
  linkcolor=darkblue,
  citecolor=darkgreen,
  urlcolor=darkblue,
  pdftitle={ESM-bench: A Benchmark for Evaluating Whether AI Agents Understand Earth System Model Physics and Code},
  pdfauthor={Koutian Wu and Yuxuan Cao and Gengchen Mai and Zeping Liu},
}

\newcommand{\esmbench}{\textsc{ESM-bench}\xspace}

\newcommand{\preprintfigpath}{figures}

\newcommand{\taskcount}{60}
\newcommand{\codebasecount}{6}

\newcommand{\pocount}{14}
\newcommand{\psscount}{13}
\newcommand{\prmcount}{16}
\newcommand{\pbfcount}{17}
\newcommand{\noahmpcount}{10}
\newcommand{\noahmppocount}{2}
\newcommand{\noahmppsscount}{3}
\newcommand{\noahmpprmcount}{2}
\newcommand{\noahmppbfcount}{3}
\newcommand{\clmfivecount}{10}
\newcommand{\clmfivepocount}{2}
\newcommand{\clmfivepsscount}{2}
\newcommand{\clmfiveprmcount}{3}
\newcommand{\clmfivepbfcount}{3}
\newcommand{\summacount}{10}
\newcommand{\summapocount}{2}
\newcommand{\summapsscount}{2}
\newcommand{\summaprmcount}{3}
\newcommand{\summapbfcount}{3}
\newcommand{\parflowcount}{10}
\newcommand{\parflowpocount}{3}
\newcommand{\parflowpsscount}{2}
\newcommand{\parflowprmcount}{2}
\newcommand{\parflowpbfcount}{3}
\newcommand{\momsixcount}{10}
\newcommand{\momsixpocount}{2}
\newcommand{\momsixpsscount}{2}
\newcommand{\momsixprmcount}{3}
\newcommand{\momsixpbfcount}{3}
\newcommand{\viccount}{10}
\newcommand{\vicpocount}{3}
\newcommand{\vicpsscount}{2}
\newcommand{\vicprmcount}{3}
\newcommand{\vicpbfcount}{2}
\newcommand{\modificationcount}{49}
\newcommand{\implementationcount}{11}
\newcommand{\referenceapplycount}{60}

\newcommand{\sourcefilesmedian}{2}
\newcommand{\sourcefilesmin}{1}
\newcommand{\sourcefilesmax}{10}
\newcommand{\changedlinesmin}{5}
\newcommand{\changedlinesmedian}{90}
\newcommand{\changedlinesmax}{497}

\title{\textbf{\esmbench: A Benchmark for Evaluating Whether AI Agents Understand Earth System Model Physics \emph{and} Code}}

\author{%
  Koutian Wu\textsuperscript{1}\thanks{Corresponding author}\thanks{\href{https://orcid.org/0000-0002-8409-3579}{ORCID: 0000-0002-8409-3579}} \quad
  Yuxuan Cao\textsuperscript{1}\thanks{\href{https://orcid.org/0000-0003-3565-906X}{ORCID: 0000-0003-3565-906X}}\\[0.25em]
  Gengchen Mai\textsuperscript{1}\thanks{\href{https://orcid.org/0000-0002-7818-7309}{ORCID: 0000-0002-7818-7309}} \quad
  Zeping Liu\textsuperscript{1}\thanks{\href{https://orcid.org/0000-0003-2898-0023}{ORCID: 0000-0003-2898-0023}}\\[0.5em]
  \textsuperscript{1}The University of Texas at Austin\\
  *Corresponding author: \texttt{ktwu@utexas.edu}
}

\begin{document}

\maketitle

\begin{abstract}
AI coding agents that modify Earth System Model (ESM) code must preserve both software interfaces and the physical assumptions encoded in them. We introduce \esmbench, a benchmark of \taskcount{} developer-commit tasks from \codebasecount{} repositories: Noah-MP, CLM5, SUMMA, ParFlow, MOM6, and VIC. The v5 corpus contains \pbfcount{} physics-based bug fixes, \prmcount{} process-representation modifications, \psscount{} parameterization-scheme selections, and \pocount{} parameter-optimization tasks. Each task includes a scientific objective, parent-commit source files, a reference diff, and three stored prompt variants. An artifact audit confirms that all \referenceapplycount{} reference diffs apply to their supplied starting files. We define an eight-item patch-review rubric and audit five archived patch fixtures against the implemented local checkers. None of the five fixture diffs applies to its supplied starting files, yet three text-pattern checkers return a pass. A ParFlow expression replay disagrees with the reference slopes, and a MOM6 pattern check detects an omitted update. Inspection also identifies a retained VIC reference assignment that overwrites the intended vegetation-specific albedo. These findings expose limits of reference patches and local proxies as correctness evidence. The release provides a reproducible corpus and diagnostic tools; complete expert validation, comparative agent performance, and full-model execution remain unmeasured on v5.

\end{abstract}

\section{Introduction}
\label{sec:intro}

Earth System Models (ESMs) represent physical processes through large scientific codebases. A change to a runoff calculation, a snow parameter, or an ocean mixing routine must respect the model's interfaces, units, and physical assumptions. A syntactically valid edit can target the wrong pathway or omit a companion update. Conversely, a patch that differs from a developer's reference can still implement an acceptable scientific solution. These properties make ESM maintenance a useful setting for studying AI code modification beyond textual similarity to a reference.

Software-engineering benchmarks such as SWE-bench~\citep{jimenez2024swebench} ground evaluation in repository tasks, while SciCode~\citep{tian2024scicode} and ScienceAgentBench~\citep{chen2024scienceagentbench} extend evaluation to scientific programming and workflows. \esmbench focuses on a narrower task: generating reviewable patches for production ESM Fortran and C code when the relevant parent-commit files are supplied. It does not test unrestricted repository navigation or autonomous scientific discovery. Its scientific motivation is the need to inspect both the code change and the physical behavior it is intended to modify~\citep{clune2011software}.

We present the v5 corpus as an explicit, auditable snapshot: \taskcount{} tasks from \codebasecount{} repositories, with ten tasks per repository and four scientific-change categories. The evidence reported here has two parts. First, we audit the stored task records, source files, and reference diffs. Second, we replay five existing patch fixtures through the repository's deterministic local checks. This scope supports claims about corpus structure and the observed behavior of those checks. It does not support a ranking of current models or a population estimate of physical failure rates.

The contributions are:
\begin{enumerate}[leftmargin=1.5em, topsep=4pt, itemsep=2pt]
\item A commit-grounded corpus spanning land-surface, hydrology, and ocean code, with scientific objectives, source snapshots, reference patches, and stored prompt variants.
\item An evaluation protocol that distinguishes patch application, physics-aware review, and task-specific numerical or source-contract checks.
\item A reproducible corpus audit and five diagnostic fixture replays, with explicit separation of current evidence from historical aggregate scores and unexecuted experimental extensions.
\end{enumerate}

The empirical audit identifies a concrete evaluation failure mode: a local source-pattern checker can accept a diff that does not apply, and reference text can contain an assignment that defeats the task's stated objective. These findings motivate explicit validity gates and independent scientific review before using the corpus for comparative agent scoring.

\section{Related Work}
\label{sec:related}

\paragraph{Benchmark discovery and broader agent evaluation.}
Benchmark Radar~\citep{wu2026benchmarkradarlivingdatabase} provides a living
database and search engine for discovering AI benchmarks and inspecting their
evaluation evidence; we used it to locate several of the benchmarks discussed
in this section. Such infrastructure helps situate domain-specific
benchmarks within the broader evaluation landscape. MatrAIx~\citep{
li2026matraixsimulatingworld83} extends evaluation to interactions with diverse
simulated users through population-scale persona agents. ASI-Bench~\citep{zhou2026asibenchdawnartificialsuperintelligence} contains 60 project-level research tasks across 11 scientific domains and progressively withdraws methodological guidance, to test how far an agent proceeds without it. Research-agent benchmarks score the replication of machine-learning papers with hierarchical rubrics (PaperBench~\citep{starace2025paperbench}), open-ended research engineering against human experts (RE-Bench~\citep{wijk2024rebench}), the implementation of research extensions (RExBench~\citep{edwards2025rexbench}), follow-up research on open problems contributed by national-laboratory scientists (Into the Unknown~\citep{kim2026intotheunknown}), and the use of graphical scientific software (OSWorld-Science~\citep{dai2026osworldscience}). These efforts address benchmark discovery, human-facing behavior, and open-ended research; \esmbench focuses on the physical validity of agent-generated edits to scientific simulation code.

\paragraph{Repository and scientific-code evaluation.}
SWE-bench~\citep{jimenez2024swebench} motivates the use of real repository changes and reproducible starting states. HumanEval~\citep{chen2021evaluating} provides a widely used basis for sample-based code-generation metrics. SciCode~\citep{tian2024scicode} and ScienceAgentBench~\citep{chen2024scienceagentbench} motivate scientific objectives and task-specific assessment. \esmbench uses developer commits to obtain concrete source changes, then exposes the affected source files as the patch-synthesis workspace. Its emphasis is the relationship between a scientific objective, an implementation change, and the evidence required to assess that change.
ScienceIDE~\citep{geng2026scienceide} turns scientific code repositories into
executable environments for agent training and evaluation, guided by
expert-defined scientific cases and acceptance criteria. It supports task
generation, execution, and scientific verification. This infrastructure is
complementary to \esmbench's domain-specific evaluation of code edits and
physical validity in Earth system models. POSEIDON~\citep{zhang2026poseidon} generates executable training environments from scientific repositories at scale, and DeployBench~\citep{wang2026deploybench} evaluates agents on setting up research artifacts from a fresh machine.
SWE-bench Science~\citep{xu2026swebenchscience} is a repository-level benchmark for scientific software engineering, with 119 tasks from 98 scientific repositories in 20 domains, with failure mechanisms that include missing scientific knowledge, surface-level repair, and incomplete repair coverage. Two benchmarks argue that passing functional tests is insufficient evidence. SWE-Gate~\citep{he2026swegate} adds constraints derived from pull-request review, and PETSCAgent-Bench~\citep{zhang2026petscagentbench} scores code built on a production HPC library for solver choice, library conventions, and performance, finding failures that occur even when the code compiles and runs. \esmbench shares this position for a different criterion: whether an edit preserves the physical assumptions of an Earth system model.
Noah-Agent~\citep{wu2025noahagent} applies a multi-expert agent framework to the automated parameterization and validation of large-scale Fortran climate models, a setting close to the Earth system codes used here.

\paragraph{Earth-science, physics-simulation, and HPC code evaluation.}
In the Earth sciences, AtmosCoder-Bench~\citep{ran2026atmoscoderbench} grades executed code for environmental-science calculations and reports that models often fail to apply known formulas and constraints consistently, or revert to canonical methods when task conditions invalidate them. TerraBench~\citep{nguyen2026terrabench} evaluates agents that reason over gridded data, imagery, and simulator outputs through scientific tools, and ClimateViz~\citep{su2025climateviz} evaluates fact verification on scientific charts. RealBench~\citep{li2026realbench} evaluates data-driven weather forecasting models against operational analyses and station observations rather than reanalysis. For physical simulation, WorldSolver~\citep{jiang2026worldsolver} asks agents to implement solvers inside fixed simulation scaffolds and checks execution, visual fidelity, and physical plausibility; GPUPhysBench~\citep{sun2026gpuphysbench} measures correctness and speed of GPU physics code; and RSI-PDEbench~\citep{lu2026rsipdebench} evaluates equation identification and solver configuration from solution data. HPC-Bench~\citep{cui2026hpcbench} evaluates kernel optimization in HPC workloads, a survey reviews language models in HPC programming~\citep{ljaljevic2026llmhpcsurvey}, and ADFD-Migrate~\citep{surana2026adfdmigrate} ports legacy Fortran repositories to Python and checks behavioral agreement with the Fortran source. These benchmarks evaluate calculations, reasoning over data, forecasts, newly written solvers or kernels, or translated code; \esmbench instead poses edits to existing production Earth system models, whose correctness depends on the physical assumptions already encoded in the surrounding code.

\paragraph{Physical-model validation.}
Scientific software testing must address numerical and scientific behavior as well as implementation errors~\citep{clune2011software}. Climate-model evaluation commonly resolves results by component and physical variable~\citep{eyring2016overview,righi2020esmvaltool}. We adopt component and task-category reporting to avoid treating heterogeneous ESM edits as a uniform population. The local checks in this paper cover selected expressions and code patterns; they complement, but do not reproduce, coupled-model validation.

\paragraph{Provenance and contamination.}
Temporal benchmark designs such as LiveCodeBench~\citep{jain2024livecodebench} address a different requirement from reproducible source snapshots. A parent commit and a reference diff establish task identity, but do not show whether a model encountered the target change during training. Commit logs from a scientific research trajectory have also been mined for reusable agent skills~\citep{wu_2026_19582289}. \esmbench records commit hashes and source contents. The selected task records lack explicit source-commit timestamps, so model-specific temporal filtering requires additional upstream-history metadata.

\section{ESM-bench Design}
\label{sec:design}

\subsection{Codebases and Task Categories}
\label{sec:codebases}

The v5 corpus samples six codebases: Noah-MP~\citep{niu2011community,wu2026planthydraulics} and CLM5~\citep{lawrence2019community} for land-surface processes, SUMMA~\citep{clark2015unified}, ParFlow~\citep{kollet2006integrated}, and VIC~\citep{liang1994simple} for hydrologic processes, and MOM6~\citep{adcroft2019gfdl} for ocean processes. WRF and E3SM are outside this snapshot and contribute neither tasks nor evaluation results.

Tasks carry four category labels. \textbf{Physics-based bug fixes (PBF)} repair behavior described as physically incorrect. \textbf{Process representation modifications (PRM)} change the implemented representation of a process. \textbf{Parameterization scheme selection (PSS)} changes a scheme or its selection logic. \textbf{Parameter optimization (PO)} changes parameters or related defaults. PO is the dataset's category name; these records are source-edit tasks and do not necessarily involve an optimization algorithm or a calibration experiment. Category boundaries and scientific descriptions are annotations to be validated, rather than correctness guarantees derived from the existence of a developer commit.

\subsection{Task Records and Construction}
\label{sec:taxonomy}

Each task records a source commit, its parent hash, the developer commit message, a scientific objective and background, physical constraints, relevant files, a reference unified diff, and complete contents of the supplied parent files. The \texttt{starting\_state} is a set of relevant files rather than a complete dependency-resolved repository checkout. The source snapshot is sufficient to check application of the reference patch; compilation may require additional files and an upstream build environment.

The records identify \texttt{deepseek-v4-pro} as the metadata generator and filter, and \texttt{openai-codex-cli-0.136.0} as the metadata validator. These fields document automated preparation. They do not establish expert approval. A retained spot-check list selects six tasks for domain review, but the audited snapshot contains no completed v5 expert-label dataset from which agreement or annotation error rates can be estimated.

\subsection{Prompt Variants}
\label{sec:methodology}

Every task stores three prompts under \texttt{level\_1}, \texttt{level\_2}, and \texttt{level\_3}; the evaluation interface calls these P1, P2, and P3. The direct-prompt runner appends the complete supplied source files and requests a complete unified diff. The intended progression adds routine or line-location hints, then more explicit reference-derived code context and coding guidance. These are oracle-assisted conditions because the hints are constructed using the target change. P1 descriptions can already name files and routines, so the tier comparison is not a clean test of unrestricted code navigation.

The stored prompt strings, rather than a reconstructed description of them, define each condition. A controlled comparison requires identical task identities and source hashes across tiers. No such complete v5 solver comparison is reported here. External-document retrieval, commit-memory retrieval, process-lineage experiments, and trajectory scoring remain extensions of the protocol, without measured results in this snapshot.

\subsection{Task-specific Physical Contracts}

Five task records include a \texttt{physical\_contract}. Their subjects are topographic slopes (ParFlow), vertical-unit rescaling (MOM6), external urban coupling and leaf-mass initialization (Noah-MP), runoff accounting (SUMMA), and fresh-snow albedo (VIC). The associated tools inspect source or patch patterns and, for four tasks, evaluate extracted expressions or hand-coded scenario calculations on configured inputs. Only the ParFlow tool evaluates expressions extracted from the candidate. The Noah-MP, SUMMA, and VIC tools select reference-like outputs after matching code patterns; the MOM6 tool is a source-pattern contract. None of these tools compiles and executes the complete patched ESM.

The broader \texttt{computational\_tests} and \texttt{physical\_tests} fields are marked \texttt{scaffold\_only} in all 60 records. The five local contracts provide narrower checks despite those placeholder full-runtime fields. A gate pass is therefore a statement about its configured checks, not a claim that every physical constraint listed in the task has been tested.

\section{Dataset Statistics and Artifact Audit}
\label{sec:stats}

Table~\ref{tab:task_distribution} reports the complete v5 corpus, not an evaluated solver sample. Its \taskcount{} records correspond to 60 distinct source commits, with ten tasks per codebase. Categories are approximately balanced across the corpus. The retained tier labels classify \modificationcount{} tasks as modifications and \implementationcount{} as implementations; these labels have not been calibrated against expert-rated difficulty.

\begin{table}[t]
\centering
\caption{V5 task counts by codebase and category. Each row contributes ten tasks to the 60-task corpus.}
\label{tab:task_distribution}
\small
\begin{tabular}{llrrrrr}
\toprule
Codebase & Component & PO & PSS & PRM & PBF & Total \\
\midrule
Noah-MP & Land & \noahmppocount & \noahmppsscount & \noahmpprmcount & \noahmppbfcount & \noahmpcount \\
CLM5 & Land & \clmfivepocount & \clmfivepsscount & \clmfiveprmcount & \clmfivepbfcount & \clmfivecount \\
SUMMA & Land hydrology & \summapocount & \summapsscount & \summaprmcount & \summapbfcount & \summacount \\
ParFlow & Hydrology & \parflowpocount & \parflowpsscount & \parflowprmcount & \parflowpbfcount & \parflowcount \\
MOM6 & Ocean & \momsixpocount & \momsixpsscount & \momsixprmcount & \momsixpbfcount & \momsixcount \\
VIC & Hydrology & \vicpocount & \vicpsscount & \vicprmcount & \vicpbfcount & \viccount \\
\midrule
Total & & \pocount & \psscount & \prmcount & \pbfcount & \taskcount \\
\bottomrule
\end{tabular}
\end{table}

We reconstruct each starting workspace from its stored file contents and run \texttt{git apply --check} on the reference diff. All \referenceapplycount{} of \taskcount{} diffs pass. This check establishes internal consistency between the stored patch and starting files. It does not verify that the files match the upstream commit, that the scientific description is accurate, or that the resulting code compiles.

Tasks supply between \sourcefilesmin{} and \sourcefilesmax{} source files (median \sourcefilesmedian{}). Reference diffs contain \changedlinesmin{}--\changedlinesmax{} added plus deleted lines (median \changedlinesmedian{}), excluding diff headers. Thus the selected corpus is not restricted to the 5--150-line filter described in earlier drafts. All tasks have three stored prompt variants. Five have local physical contracts. None has an \texttt{answer\_type} label or an explicit source-commit date field, and all 60 full-runtime test records remain scaffolds. Table~\ref{tab:artifact_coverage} separates these evidence levels.

\begin{table}[t]
\centering
\caption{Observed artifact coverage in the selected v5 snapshot. Counts describe available records or completed checks, not model performance.}
\label{tab:artifact_coverage}
\small
\begin{tabularx}{\textwidth}{Xr}
\toprule
Artifact or check & Tasks \\
\midrule
Distinct tasks with source and parent hashes & 60/60 \\
Reference diff applies to supplied parent files & 60/60 \\
Three stored prompt variants & 60/60 \\
Task-specific local physical contract & 5/60 \\
Non-scaffold full-runtime test metadata & 0/60 \\
Explicit answer-multiplicity annotation & 0/60 \\
Explicit source-commit date field & 0/60 \\
\bottomrule
\end{tabularx}
\end{table}

\section{Evaluation Framework}
\label{sec:eval}

\subsection{Separate Evidence for Separate Questions}

The framework distinguishes three questions: does the patch apply, does its implementation appear scientifically appropriate under review, and does it satisfy an executable or static task-specific check? A successful check at one level does not imply success at the others. In particular, parsing a unified diff is weaker than applying it, and applying it is weaker than building the modified program.

\paragraph{Patch application.}
For task $\tau$ and submitted diff $\hat\delta$, define $V(\tau,\hat\delta)=1$ when the diff is well formed and applies to the supplied starting files. The corpus audit applies this test to all 60 reference patches and to the five retained fixtures, reporting those denominators separately. It does not report a comparative solver ValidDiff rate.

\paragraph{Physics-aware review.}
The configured v2 rubric has eight items scored 0 (fail), 1 (partial), or 2 (clear pass). Four inspect code: correct target, local consistency, compilability, and completeness. Four inspect physics: physical intent, dimensional consistency, conservation awareness, and parameterization validity. ``Compilability'' is a static plausibility judgment in this rubric, not a recorded compiler result. Conservation awareness is similarly a review of the visible change; global conservation can require execution beyond the local patch.

For a completed review with scores $r_1,\ldots,r_8$, the normalized score and review-pass predicate are
\begin{equation}
R = \frac{1}{16}\sum_{i=1}^{8}r_i,
\qquad A = \mathbf{1}\!\left[\min_i r_i\geq1\right].
\end{equation}
Code and physics sub-scores can be reported separately on their native 0--8 scales. These are protocol definitions. The present evidence does not supply a complete, expert-adjudicated v5 review table or a calibrated relationship between these scores and physical correctness. Historical six-item and current eight-item rubric outputs must not be pooled without preserving their scoring versions.

\paragraph{Reference similarity.}
Exact match and structural diff similarity measure agreement with one developer patch. Scientifically acceptable alternatives can differ structurally, so similarity is a diagnostic rather than a physical-correctness criterion. Because v5 lacks answer-multiplicity labels, this paper does not report a single-answer exact-match result.

\paragraph{Local contract gates.}
The fixture replay runs the configured gate for each available archived patch. Gate definitions mix source-pattern requirements and, for ParFlow, numerical agreement of extracted expressions with a reference kernel. The other numerical scenario outputs are conditional illustrations selected by recognizers, rather than execution of candidate code. Numerical thresholds are the fixed tolerances in the retained configuration, not estimates from repeated compiler or discretization perturbations. Section~\ref{sec:experiments} reports each gate separately so that a source-pattern failure is not confused with a measured simulation divergence.

\subsection{Reporting and Scope}

A comparative solver study should identify the task snapshot, exact solver and judge configurations, prompt content, sampling procedure, and scored-task denominators. Rubric assessment additionally requires calibration against domain experts and vendor-independent judges. Repeated sampling can support pass@$k$ estimates~\citep{chen2021evaluating}; no multi-sample solver results are reported here.

The local fixture sample is deliberately selected and small. Its outcomes cannot estimate success rates for the 60-task corpus, the omitted codebase (CLM5), or a model family. We report the count of configured gates passed as a fixture-replay diagnostic, without a population confidence interval or leaderboard interpretation.

\section{Results: Reproducible Local Fixture Checks}
\label{sec:experiments}

\subsection{Evidence Snapshot and Replay Setup}

The current results comprise the 60-task artifact audit and a replay of five archived patch fixtures on 6 September 2026. The fixtures carry the identifier \texttt{codex\_solver}; a pinned solver model ID, generation prompt, token usage, and per-call cost are not retained with them. We make no attribution to a particular Codex or foundation-model version. No new solver or judge calls were made for this replay.

The command \texttt{scripts/run\_sim\_compare.py} uses \texttt{configs/simulation\_compare.yaml} to select five task records, their archived diffs, and registered local checks. One checker evaluates extracted ParFlow expressions; the other four recognize source patterns. Three recognizers additionally calculate illustrative scenario outputs from the recognized branch. Native model compilation and full ESM execution are outside the replay. The five tasks comprise two PBF tasks and three PO tasks, with one fixture from each of five repositories. They are not a random sample. Table~\ref{tab:gate_results} states the evidence for each outcome.

\begin{table}[t]
\centering
\caption{Replay of five archived fixtures. A pass means that the configured local checker returns true; all five fixture diffs fail application checks. These results do not measure full-model behavior or solver success on v5.}
\label{tab:gate_results}
\small
\setlength{\tabcolsep}{4pt}
\begin{tabularx}{\textwidth}{l l X}
\toprule
Task & Gate & Observed evidence \\
\midrule
PF-PO-001 & Fail & Extracted slope expressions differ from the reference on a $512\times512$ grid; RMSE is 0.1131 for $x$ and 0.0790 for $y$. \\
MOM6-PO-007 & Fail & Four of five required source-pattern checks pass; the bulk mixed-layer friction-velocity default update is absent. \\
NOAH-PBF-003 & Pass & Five source-pattern checks pass. Reported leaf-mass agreement is conditional on recognized patterns, not candidate execution. \\
SUMMA-PBF-001 & Pass & Four source-pattern checks pass. The checker selects the reference runoff expression when those patterns match. \\
VIC-PO-001 & Pass & Five source-pattern checks pass. The checker assigns reference albedos to the candidate branch when patterns match. \\
\bottomrule
\end{tabularx}
\end{table}

\subsection{Reference Disagreement in the ParFlow Fixture}

\begin{figure}[t]
\centering
\includegraphics[width=\textwidth]{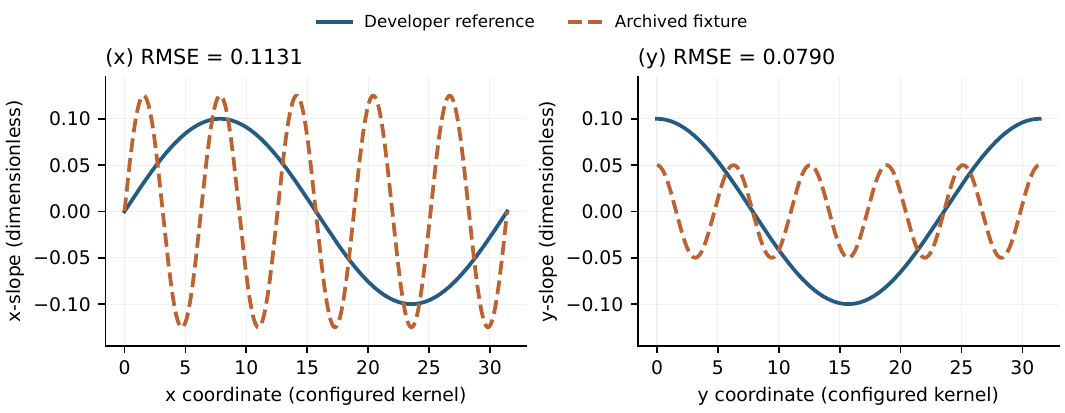}
\caption{One-dimensional slices of the slope expressions evaluated in the ParFlow fixture replay. Reference and fixture curves are reconstructed from the archived result artifact on the configured coordinates. The difference in wavelength persists even though both fields remain finite. Slopes are dimensionless; coordinate units follow the configured kernel.}
\label{fig:slope_replay}
\end{figure}

The ParFlow check evaluates the active slope assignments extracted from the parent source, developer diff, and archived fixture. The reference uses $0.10\sin(x/5)$ and $0.10\cos(y/5)$; the fixture uses $\sin(x)/8$ and $\cos(y)/20$. Both fixture fields are finite on the configured grid, but their wavelength and amplitude differ from the reference (Figure~\ref{fig:slope_replay}). Maximum absolute differences are 0.2064 and 0.1322, respectively. Both exceed the configured $10^{-5}$ maximum-error tolerance; their RMSE values exceed the $10^{-6}$ tolerance.

This is a reproducible mismatch with the reference kernel. It does not demonstrate that every alternative periodic slope field is physically invalid. Determining whether an alternative is acceptable requires a sufficiently specific scientific objective and independent review of the contract. The gate currently enforces reference agreement on the selected expressions.

\subsection{Patch Application and Recognizer Outcomes}

All five archived fixture diffs fail \texttt{git apply --check} against their corresponding v5 parent files. Git reports corrupt patches for MOM6, Noah-MP, SUMMA, and VIC; the ParFlow hunks fail to match the supplied source. These failures are distinct from the 60 successful reference-patch checks. The local-check runner reads diff text without enforcing application, so its outputs describe text-derived diagnostics on invalid fixtures rather than validated submitted programs.

The MOM6 recognizer finds four of five required update patterns, omitting the bulk mixed-layer friction-velocity default. It searches diff-hunk text, including removed lines, which limits the interpretation of even a matched pattern. Noah-MP, SUMMA, and VIC return local-check passes. Their zero-error scenario results are conditional on regex recognition: for example, the SUMMA tool sets candidate runoff equal to the hand-coded reference calculation if its source patterns match, and VIC assigns the reference albedo dictionary to the candidate branch. Such agreement is built into the checker and supplies no independent numerical validation of candidate code.

Thus three of five local checkers accept their fixture text, while zero of five fixtures satisfy patch application. This is an audit of evaluation instrumentation, not a model success estimate. The configuration additionally sets \texttt{judge\_passed=true} and \texttt{plausible\_patch=true}; these are supplied flags rather than observed review outcomes. The derived \texttt{silent\_physics\_failure} labels are not used as scientific results.

\subsection{A Reference-Patch Consistency Issue}

The retained VIC-PO-001 developer diff adds a branch that assigns vegetation-specific fresh-snow albedo, but the following unconditional assignment restores the global fresh-snow value:
\begin{lstlisting}[language=C]
if (options.MAX_SNOW_ALBEDO) {
    if (iveg != Nveg) {
        snow->albedo = new_snow_albedo;
    }
    else {
        snow->albedo = param.SNOW_NEW_SNOW_ALB;
    }
}
snow->last_snow = 0;
snow->albedo = param.SNOW_NEW_SNOW_ALB;
\end{lstlisting}
The later assignment overwrites the earlier value in this branch. The stored reference therefore does not establish the vegetation-specific behavior described by the local contract. This source-level finding illustrates why developer provenance and patch application alone cannot certify scientific correctness. We retain the original record for auditability and flag the task as unsuitable for correctness scoring until its objective and reference are adjudicated. The corpus statistics include this flagged record; they are counts of released task records, not of expert-certified tasks.

\section{Analysis and Limitations}
\label{sec:analysis}

\paragraph{What the current evidence establishes.}
The corpus is internally consistent at the reference-patch application level: all 60 reference patches apply to their supplied parent files. The fixture replay exposes disagreement between patch application and local recognizer acceptance, and quantifies a ParFlow expression mismatch. These are useful reproducibility results, but they do not establish benchmark difficulty, calibrated scientific correctness, or comparative model capability.

\paragraph{Expert validation and annotation.}
The scientific descriptions, category labels, and physical contracts need completed expert adjudication. The snapshot includes automated validator metadata and a review-selection list, not a measured human agreement study. There is no supported Cohen's kappa, category confusion matrix, or annotation error rate to report. The lack of answer-multiplicity labels also prevents a defensible single-answer exact-match subset. The VIC overwrite finding is a concrete reason to adjudicate objectives and references before treating the corpus as a scored benchmark.

\paragraph{Contract coverage.}
The five gates use task-specific recognizers and simplified calculations. Only ParFlow independently evaluates expressions extracted from candidate text; three other tools select reference outputs after pattern recognition. A recognizer can miss a valid alternative implementation, and a passing recognized pattern can coexist with unrelated errors elsewhere in the patch. Finite outputs and reference agreement on a few scenarios do not establish long-term water, energy, or tracer conservation. Complete builds, broader tests, and scientific review are necessary to assess those properties.

\paragraph{Sampling and comparisons.}
The six repositories each contribute ten selected tasks; this design does not represent the natural frequency of scientific code changes. The five fixtures omit CLM5 and both PRM and PSS categories. Their solver identity and generation configuration are incomplete. Consequently, neither between-model comparisons nor effects of prompt variants, retrieval, or agent scaffolds can be inferred from these results.

\paragraph{Public-history exposure.}
The commits and source code are public. Hashes make a task identifiable, but do not establish training-data exclusion. Timestamp metadata and an audited temporal split are needed before making freshness or memorization claims. Stored memory-exclusion policies express intended rules, not evidence that a particular retrieval run followed them.

\paragraph{Cost and reproducibility.}
The local replay makes no paid model calls. Historical solver and judge costs, tokens, provider regions, and exact call configurations are not recoverable from the retained fixtures, so cost-efficiency comparisons are unsupported. Future measured evaluations need to retain that information alongside per-task outputs. Reproducibility here means regenerating corpus statistics and local gate outcomes from archived inputs, rather than reproducing unspecified model calls.

\section{Discussion and Conclusion}
\label{sec:conclusion}

\esmbench provides a 60-task v5 corpus for physics-aware code modification across six ESM repositories. The tasks combine developer-commit provenance, supplied source files, reference patches, scientific descriptions, and prompt variants. The artifact audit verifies reference-patch application for every task. A five-fixture audit reveals that local recognizers can accept non-applying diffs. A ParFlow expression replay quantifies reference disagreement, while a VIC source inspection identifies a reference assignment that defeats the stated albedo objective.

The current evidence supports these corpus and tool results. It does not support claims that a particular model understands ESM physics, that one model outperforms another, or that successful review systematically hides physical failures. A complete v5 evaluation requires traceable solver outputs, calibrated expert review, matched prompt conditions, and broader executable coverage. The released protocol makes those requirements explicit while preserving the narrower results that can already be reproduced.

\section*{Acknowledgments}
The authors thank Zesen Huang and Hejia Geng for community development and collaboration, as well as discussions and feedback on scientific benchmarking.

We thank Tian Zhou, Lingcheng Li, Weihao Liu, Mashrekur Rahman, Yuhan Wang, Cenlin He, Harsh Kamath, participants at HydroML 2026 and the 4th Annual Good Systems Smart Cities and AI Innovations Symposium, and other researchers and collaborators for discussions and feedback related to Earth system modeling, benchmark design, and physical evaluation.

We thank the developers of Noah-MP, CLM5, SUMMA, ParFlow, VIC, and MOM6 for openly developing these models; the benchmark tasks are built from their public commit histories.

The authors used AI tools to assist with software development and editing, and take responsibility for the content of this paper.

This project was supported by NSF NCAR/CISL Allocation Project UTAA0012. Some simulations were performed on the NCAR CISL and Texas Advanced Computing Center (TACC) supercomputer systems.

\bibliographystyle{unsrtnat}
\bibliography{references}

\appendix
\section{Reproducibility and Evidence Inventory}
\label{app:construction}

\subsection{Selected Snapshot}
\label{app:reproducibility}

The selected corpus is \texttt{data/tasks\_v5}. The paper's \texttt{preprint\_snapshot.json} lists all 60 task paths, task and source-commit identities, SHA-256 file hashes, category and tier metadata, source-file counts, reference-diff application results, and contract coverage. Its manifest hash identifies the task file contents rather than relying on a moving directory name. The paper uses \texttt{preprint\_metrics.tex}, generated from this explicit corpus. Earlier \texttt{generated\_metrics.tex} remains a historical artifact and is not included in this preprint's numerical results.

The corpus audit can be rerun from the repository root with:
\begin{lstlisting}
python3 scripts/generate_preprint_snapshot.py
\end{lstlisting}
The script reconstructs each supplied starting workspace in a temporary directory and checks the reference patch with \texttt{git apply --check}. It excludes difficulty sidecars from task counts. It does not download upstream repositories or query model APIs. The check verifies stored-source consistency, not external provenance against a fresh upstream checkout.

\subsection{Local Gate Replay}
\label{app:executable_audit}

The replay uses Python 3.10 or newer, NumPy, and PyYAML. It can be regenerated with:
\begin{lstlisting}
python3 scripts/run_sim_compare.py \
  --config configs/simulation_compare.yaml \
  --output-dir data/runs/sim_compare/preprint_validation_20260906
\end{lstlisting}
The output directory must be inside the repository because the runner records a repository-relative path. The archived replay contains the resolved YAML configuration, a summary, and five per-task result files. The input diffs are in \texttt{data/sim\_compare\_fixtures}. The replay was run on 6 September 2026 using evaluation code from repository commit \texttt{30c4d76}; the preprint revision changes documentation and adds audit/figure tooling, without changing that gate implementation.

The ParFlow kernel samples 512 points per coordinate over $x\in[0.001,10\pi]$ and $y\in[0,10\pi]$. RMSE and maximum-error tolerances are $10^{-6}$ and $10^{-5}$. The MOM6 gate checks five source patterns and has no numerical trajectory comparison. Noah-MP compares leaf mass to a $10^{-12}$~g\,m$^{-2}$ absolute tolerance in two scenarios. SUMMA uses a $10^{-12}$~m\,s$^{-1}$ tolerance on runoff comparisons in two scenarios. VIC uses a $10^{-12}$ albedo tolerance and a $10^{-9}$~W\,m$^{-2}$ absorbed-shortwave tolerance in two scenarios. These are configured engineering thresholds; no tolerance-calibration experiment is claimed.

\subsection{Interpreting Checker Outputs}
\label{app:evaluation_details}

The archived field \texttt{sim\_pass} means the registered local checker accepted the fixture text. Patch application is checked separately by the corpus-audit script. Only ParFlow evaluates numerical expressions extracted from the fixture. MOM6 recognizes diff patterns, including removed hunk lines. Noah-MP, SUMMA, and VIC compute illustrative scenario outputs after recognizing expected text, so their zero differences from hand-coded references are not independent execution results. The archive preserves these raw outputs, but they must be read with these implementation limits.

The task manifest includes reference-diff application results and separate fixture-diff failures. The VIC-PO-001 overwrite identified in Section~\ref{sec:experiments} is a qualitative source audit, not a complete estimate of reference defects across the corpus. The task remains in the released record set with this explicit caveat.

\subsection{Requirements for a Comparative Evaluation}
\label{app:evaluation_infrastructure}

A model comparison on this corpus needs a task manifest and content hashes; the exact stored prompts and appended source; pinned model identifiers and provider configuration; sampling parameters and timestamps; complete responses, extracted diffs, and application checks; per-item review scores with rubric and judge versions; and token, latency, and cost records. Human calibration should report its sampling procedure and actual expert agreement. Prompt comparisons need matched tasks. Any claim about full-model execution needs reproducible builds and runtime outputs beyond local pattern checks.

The current code includes an eight-item review rubric and review-packet tooling. Their availability is not evidence of completed human validation. Context retrieval, commit-derived memory, trajectory scoring, and long-horizon physical audits are not evaluated in this preprint.

\section{Data and Code Availability}

The v5 task records, fixture diffs, audit and local-check scripts, configurations, and archived outputs needed to reproduce the reported checks are retained by the authors and are not distributed with this version. No public code or dataset deposit is claimed in this version. Historical model-run aggregates are excluded from the reported results because their underlying solver and judge artifacts are not included in the release.

\section{Broader Impact}

The benchmark can help make review of AI-generated scientific code more explicit. Its main risk is overinterpreting a local score as evidence that a patch is safe for production modeling. Neither reference-patch agreement nor success on the selected kernels replaces domain review and testing in the full scientific workflow. All task records derive from public repository histories; users of the benchmark should retain the upstream source attribution and applicable licenses.

\end{document}